\def\year{2021}\relax
\documentclass[letterpaper]{article} 
\usepackage{aaai21}  
\usepackage{times}  
\usepackage{helvet} 
\usepackage{courier}  
\usepackage[hyphens]{url}  
\usepackage{graphicx} 
\usepackage{natbib}  
\usepackage{caption} 
\usepackage{times}
\usepackage{epsfig}
\usepackage{graphicx}
\usepackage{amsmath}
\usepackage{amssymb}
\usepackage{subfigure}
\usepackage{tabularx}
\usepackage{color}
\usepackage{multirow}
\usepackage{rotating}
\usepackage{diagbox}
\usepackage{bm}
\usepackage{caption}
\makeatletter 
\renewcommand{\@thesubfigure}{\hskip\subfiglabelskip}
 \makeatother
\usepackage[switch]{lineno}

\title{EMLight: Lighting Estimation via Spherical Distribution Approximation}
\author{Anonymous AAAI Submission\\Paper ID: 4302} 

\author {
    Fangneng Zhan \thanks{equal contribution} \textsuperscript{\rm 1},
    Changgong Zhang \footnotemark[1] \textsuperscript{\rm 2},
    Yingchen Yu \textsuperscript{\rm 1},
    Yuan Chang \textsuperscript{\rm 3}, \\
    Shijian Lu \thanks{corresponding author} \textsuperscript{\rm 1},
    Feiying Ma \textsuperscript{\rm 2},
    Xuansong Xie \textsuperscript{\rm 2}
    \\
}
\affiliations {
    \textsuperscript{\rm 1} Nanyang Technological University,
    \textsuperscript{\rm 2} Alibaba DAMO Academy \\
    \textsuperscript{\rm 3} Beijing University of Posts and Telecommunications \\
    \{fnzhan,shijian.lu\}@ntu.edu.sg, yingchen001@e.ntu.edu.sg, changyuan@bupt.edu.cn, \{changgong.zcg,feiying.mfy\}@alibaba-inc.com, xingtong.xxs@taobao.com
}

\begin{document}


\maketitle

\begin{abstract}
Illumination estimation from a single image is critical in 3D rendering and it has been investigated extensively in the computer vision and computer graphic research community. On the other hand, existing works estimate illumination by either regressing light parameters or generating illumination maps that are often hard to optimize or tend to produce inaccurate predictions. We propose Earth Mover’s Light (EMLight), an illumination estimation framework that leverages a regression network and a neural projector for accurate illumination estimation. We decompose the illumination map into spherical light distribution, light intensity and the ambient term, and define the illumination estimation as a parameter regression task for the three illumination components. Motivated by the Earth Mover's distance, we design a novel spherical mover's loss that guides to regress light distribution parameters accurately by taking advantage of the subtleties of spherical distribution. Under the guidance of the predicted spherical distribution, light intensity and ambient term, the neural projector synthesizes panoramic illumination maps with realistic light frequency. Extensive experiments show that EMLight achieves accurate illumination estimation and the generated relighting in 3D object embedding exhibits superior plausibility and fidelity as compared with state-of-the-art methods.
\end{abstract}

\section{Introduction}
Lighting estimation aims to recover illumination from a single image with limited field of view. It has a wide range of applications in various computer vision and computer graphics tasks such as high-dynamic-range (HDR) relighting in mixed reality, etc. However, lighting estimation is an under-constrained problem as it aims to recover a 360-degree full-view illumination map from an image with limited field of view. In addition, high-dynamic-range (HDR) illumination is required to be inferred from low-dynamic-range (LDR) observations for the purpose of realistic object relighting.

Lighting estimation has been tackled through direct generation of illumination maps \cite{gardner2017,song2019,srinivasan2019lighthouse} or regression of parameters of representative illumination functions such as spherical harmonics function \cite{cheng2018shlight,garon2019fast} and spherical Gaussian function \cite{gardner2019deeppara, li2020rendering}.
However, the functional representation methods struggles to regress accurate frequency information (especially high-frequency information) that often leads to inaccurate shading and shadow effects in relighting \cite{garon2019fast} or require complex optimization steps \cite{gardner2019deeppara}. Directly generating illumination maps can preserve some high-frequency information, but it can hardly recover other information of the light sources such as light directions and sizes \cite{chen2019neural}.

\begin{figure}[t]
\centering
\includegraphics[width=1.0\linewidth]{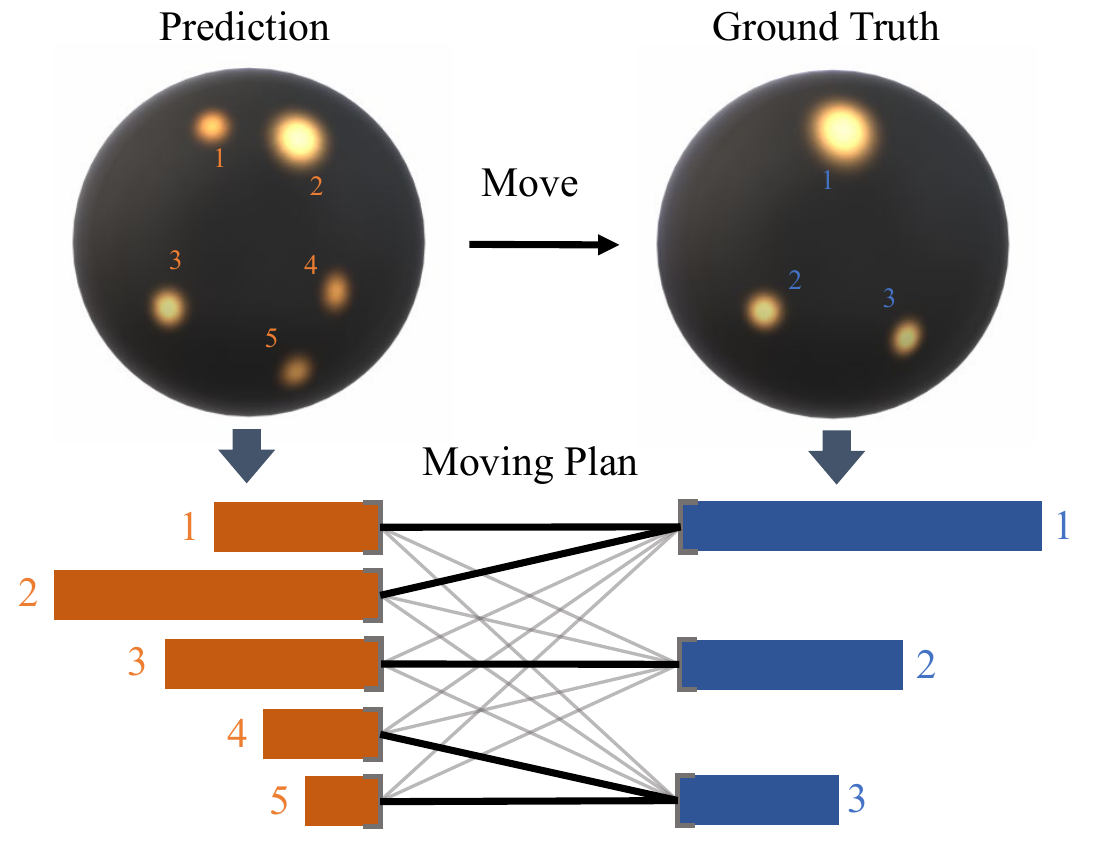}
\caption{Illustration of our proposed Earth Mover’s Light (EMLight): EMLight treats two illumination maps as two discrete spherical distributions. Motivated by the idea of Earth Mover's distances, we design a spherical mover's loss (SML) to measure the distance between two spherical distributions by calculating the minimum distance of moving one distribution to another along the spherical surface. 
SML aims to find the best \textit{Moving Plan} (with minimum total distance) as illustrated by connections between the two distributions. 
The thickness of connecting lines denotes the amount of `Earth' moved between the two points.
}
\label{im_intro}
\end{figure}

In this work, we propose \textbf{EMLight} (\textbf{E}arth \textbf{M}over's \textbf{Light}), an accurate illumination estimation framework that is capable of locating light sources and recover illumination with realistic frequency simultaneously. EMLight consists of an inter-connected regression network and neural projector, where the regression network predicts illumination parameters accurately and the neural projector leverages the estimated illumination parameters to synthesize illumination maps with realistic frequency information. Instead of regressing illumination parameters separately without considering them as a whole as in many existing works, we formulate the overall scene illumination by a spherical distribution and treat the illumination estimation as the regression of a spherical distribution as illustrated in Fig. \ref{im_intro}.

For accurate illumination representation, we decompose the illumination map into \textit{light distribution}, \textit{light intensity}, and \textit{ambient term} that capture the energy distribution of light sources, the overall intensity of light sources, and the average of remaining energy excluding light sources, respectively.
As illumination maps are spherical images,
we define $N$ anchor points on a unit sphere to model discrete light distributions. 
The task of illumination prediction is thus converted to the problem of regressing light distribution, light intensity and ambient term.
The light intensity and ambient term are scalar values, which can be directly regressed with a naive L2 loss by the regression network.
However, directly regressing $N$ discrete values (at $N$ anchor points) of light distribution with a naive L2 loss or cross-entropy loss is undesirable as this does not take advantage of the subtleties of spherical distributions such as the spatial information. 

Inspired by the Earth Mover's distance \cite{emd} that measures the distance between two distributions, we design a spherical mover's loss to regress light distributions by conducting `Earth Mover' on the unit sphere as illustrated in Fig. \ref{im_intro}. SML evaluates the distance between two spherical distributions by measuring the minimum radian distance required to move one spherical distribution to another along the spherical surface, and the target is to find the optimal moving plan among all possible moves between two distributions. It captures spatial information of spherical distribution, which greatly helps for accurate estimation of spherical light distribution.

Under the guidance of illumination parameters that are predicted by the regression network, the neural projector generates accurate illumination maps with realistic frequency information in an adversarial manner. Different from normal images, the illumination map is a panorama that usually suffers from different levels of spherical distortions at different latitudes. We therefore adopt spherical convolution \cite{spherenet} for the accurate generation of panoramic illumination maps.

The contribution of this work can be summarized in three aspects. First, we formulate the illumination estimation as the regression of the spherical distribution of illumination. To the best of our knowledge, this is the first work that estimates illumination in this manner. 
Second, we design a novel spherical mover's loss that takes advantage of the subtleties of spherical illumination distributions.
Third, we design a neural projector that employs spherical convolution to synthesize panoramic illumination maps with realistic frequency information through adversarial training.

\begin{figure*}[t]
\centering
\includegraphics[width=1.0\linewidth]{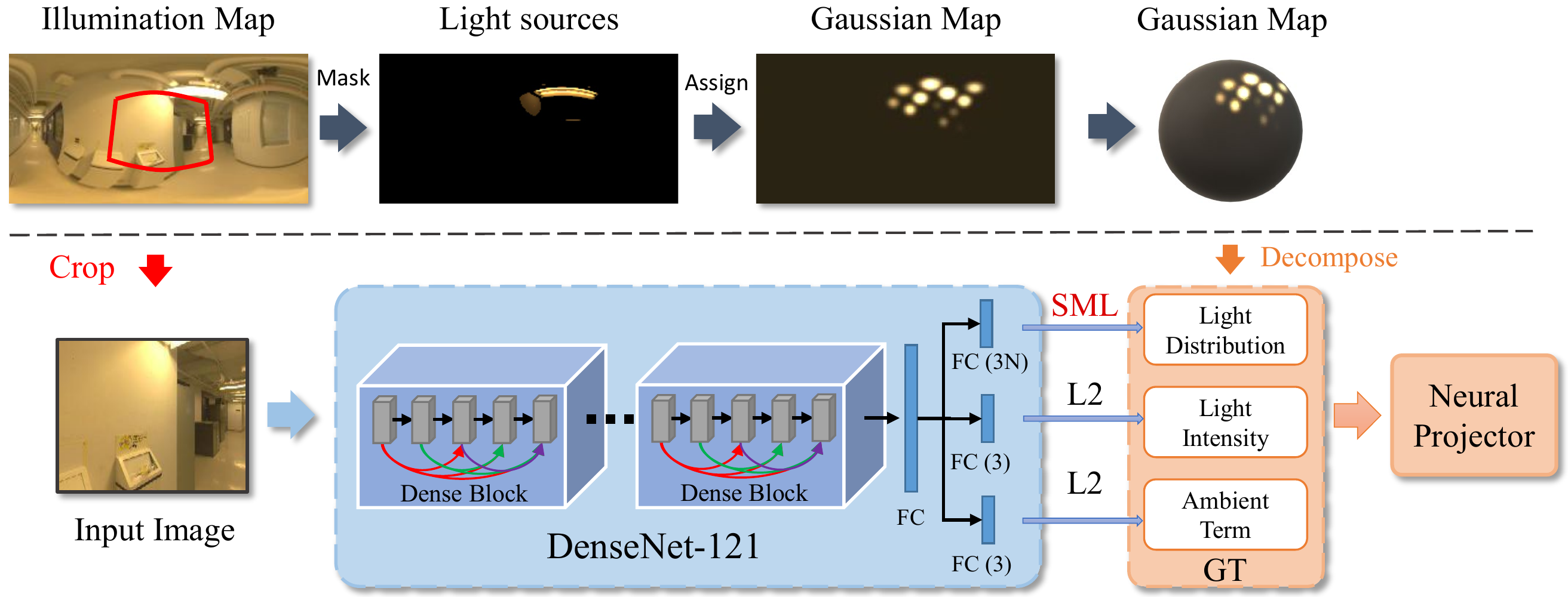}
\caption{
Illumination decomposition and estimation in EMLight: The upper and lower graphs illustrate the illumination map decomposition and the structure of the light parameter regression network, respectively. Given an \textit{Illumination Map}, we first derive the \textit{Light sources} region via thresholding and then assign light source pixels to $N$ anchor points as illustrated in \textit{Gaussian Map} (visualized by spherical Gaussian function). We decompose the illumination map into light distribution, light intensity and ambient term and use the decomposition as ground truth for regression network training. The regression network employs a local region (as highlighted by the red box) as the input and three fully-connected layers (FC) with output size of $3N$, $3$ and $3$ (RGB images have 3 channels) to regress the light distribution, light intensity and ambient term, respectively. The estimated illumination parameters are fed to the \textit{Neural Projector} for illumination map generation.
}
\label{im_stru1}
\end{figure*}

\section{Related Works}
Lighting estimation is a classic challenge in computer vision and computer graphics, and it is critical for realistic relighting in objects insertion and image synthesis \cite{lalonde2012,barron2015,geoffroy2017,murmann2019dataset,zhan2019sfgan,zhan2018verisimilar,zhan2019esir,zhan2019gadan,zhan2019acgan,zhan2019scene,zhan2020towards,zhan2020sagan,boss2020,xue2018accurate}. Traditional approaches require user intervention or assumptions about the underlying illumination model, scene geometry, etc. For example, \citet{karsch2011} recover parametric 3D lighting from a single image but requires user annotations for initial lighting and geometry estimates. \citet{zhang2016} require a full multi-view 3D reconstruction of scenes. \citet{lombardi2016} estimate illumination from an object of known shape with a low-dimensional model. 
\cite{maier2017} makes use of additional depth information to recover spherical harmonics illumination. 

On the other hand, the recent works aim to estimate lighting from images by regressing representation parameters \cite{cheng2018shlight,gardner2019deeppara,li2020rendering} or generating illumination maps \cite{gardner2017,song2019}.
\citet{garon2019fast} estimate lighting by predicting spherical harmonic (SH) coefficients from a background image and local patch. 
\citet{gardner2019deeppara} estimate the positions, intensities, and colours of light sources and reconstructs illumination maps with a spherical Gaussian function. 
On top of it, \citet{li2019spherical} represent illumination maps with multiple spherical Gaussian functions and regresses the corresponding Gaussian parameters for lighting estimation. 
\citet{gardner2017} generate illumination maps directly with a two-steps training strategy. 
\citet{song2019} estimate per-pixel 3D geometry and uses a convolutional network to predict unobserved contents in the environment map. 
\citet{legendre2019deeplight} regress HDR lighting from LDR images by comparing the rendered sphere with predicted illumination to the ground truth.
\citet{srinivasan2019lighthouse} estimate a 3D volumetric RGB model of a scene and uses standard volume rendering to estimate incident illuminations. 
Given any illumination map, the framework proposed by \citet{sun2019} is able to achieve relighting on the RGB portrait image taken in an unconstrained environment.
Besides, several works \cite{liu2020shadow,zhan2020aicnet} adopt Generative Adversarial Network to generate shadow without explicitly estimating the illumination map.

The aforementioned works either lose realistic frequency information or produce inaccurate light sources in illumination estimation. We decompose illumination maps into three components and design a regression network with a spherical mover's loss that estimates the decomposed illumination parameters with accurate spatial information. Using the estimated illumination parameters as guidance, our designed neural projector generates illumination maps accurately with realistic frequency information. 

\section{Proposed Method}

\begin{figure*}[t]
\centering
\includegraphics[width=1.0\linewidth]{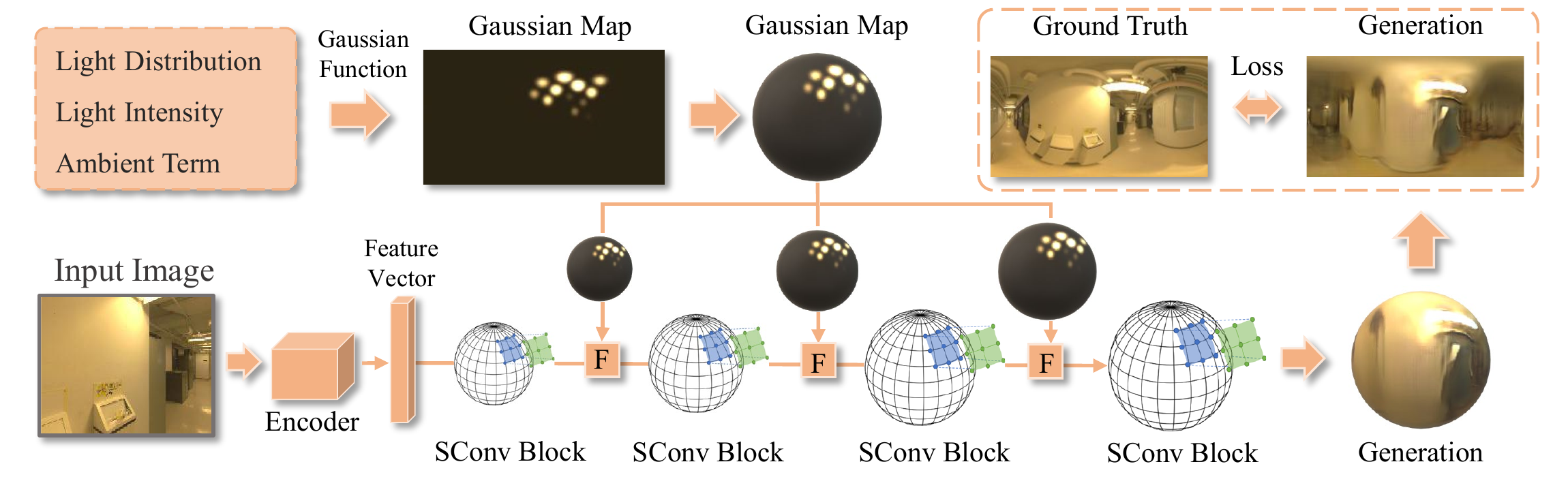}
\caption{
The structure of neural projector: \textit{SConv Block} denote spherical convolution block, \textit{F} denotes feature fusion blocks, and \textit{Gaussian Map} is acquired through spherical Gaussian mapping according to the predicted parameters. The \textit{Input Image} is fed to an \textit{Encoder} to produce a feature vector for the ensuing spherical generation. The \textit{Gaussian Map} is fused with the multi-scale spherical generation to synthesize the final illumination map.
}
\label{im_stru2}
\end{figure*}

EMLight consists of two sequential modules including a regression network and a neural projector as illustrated in Figs. \ref{im_stru1} and \ref{im_stru2}. The illumination parameters estimated by the regression network will guide the neural projector to generate illumination maps accurately. 

\subsection{Regression Network}

The structure of the regression network is shown in Fig. \ref{im_stru1}. The regression network aims to estimate three set of our decomposed illumination parameters including \textit{\textbf{light distribution $P$}}, \textit{\textbf{light intensity $I$}} and \textit{\textbf{ambient term $A$}}, which will be explained as below.
For clarity, we take one channel of RGB images as an example in the following description. We first separate light sources from illumination maps since light sources in scenes are most critical in illumination prediction. Following \cite{gardner2019deeppara}, we separate the light source by taking the 5\% pixels that have the highest values within the HDR illumination map. The \textit{\textbf{Light intensity $I$}} can then be determined by the summation of all the pixels within the light sources region, and the the \textit{\textbf{ambient term $A$}} is further determined by the averaged pixel value within the remaining region (excluding light sources). We then employ Vogel's method \cite{vogel1979} to generate $N$ (N=128 by default in this work) uniformly distributed anchor points on a unit sphere. The value of pixels within the light source region will be assigned to the anchor point with minimum radian distance. The value of each anchor point is the summation of all assigned pixel values. The value of all anchor points will be normalized by the intensity $I$ to ensure their summation equals one, so that the N anchor points form a standard discrete distribution on a unit sphere as denoted by \textit{\textbf{light distribution $P$}}. 

Three branches as shown in Fig. \ref{im_stru1} are adopted to regress the three sets of parameters respectively.
For the light intensity $I$ and ambient term $A$, a naive L2 loss can be adopted for the regression.
But for the light distribution $P$ which are localized on a sphere, a naive L2 loss cannot effectively utilize the spatial information of spherical distribution and the property of standard distribution (the summation of all anchor point values equals one).
We take advantage of the subtleties of spherical distribution and propose a novel spherical mover's loss to regress light distribution.

\subsection{Spherical Mover's Loss}

A naive method to predict the discrete spherical distribution is using L2 loss or cross-entropy loss to regress the values of N anchor points, but this naive method often introduce various problems.
Firstly, L2 loss only regresses each anchor point separately and cannot effectively evaluate the discrepancy between two sets of anchor points (two distributions).
Secondly, both L2 loss and cross-entropy loss cannot effectively utilize the spatial information of the discrete light distribution which is localized on the spherical surface.

Inspired by the Earth Mover's distance which measures the discrepancy between two distributions, we propose a novel spherical mover's loss (SML) to measure the discrepancy between two discrete spherical distributions.
To derive the SML, we define two discrete distributions with $N$ points on the sphere as denoted by $U$ and $V$.
Intuitively, SML can be treated as the minimum amount of work required to transform $U$ into $V$, where the work is measured by multiplying the amount of distribution to be moved and the distance to be moved.
Then a transportation plan (or moving plan) matrix $T$ with size of $(N, N)$ can be defined, where each entry $T_{ij}$ in $T$ represents the the amount of probability moved between point $U_{i}$ and point $V_j$.
Besides, a cost matrix $C$ with size of $(N, N)$ is also defined where each entry $C_{ij}$ in $C$ gives the distance of moving $U_{i}$ to $V_{j}$.
Specifically, the distance between a point $U_{i}$ and another point $V_{j}$ is measured by their radian distance along the unit sphere.
As the $N$ anchor points on the sphere surface are pre-defined by the Vogel's method \cite{vogel1979}, the cost matrix $C$ can be easily pre-computed as a constant matrix in training.
With the defined transportation plan matrix $T$ and cost matrix $C$, SML can be formulated as follows:
\begin{equation}
\begin{split}
& L_{sml} = \mathop{min}\limits_{T} (\sum_{i=1}^{N} \sum_{j=1}^{N} C_{ij} T_{ij}) = \mathop{min}\limits_{T} \langle C, T \rangle \\
& subject \ to \quad T\cdot \vec{1} = U, \quad T^\top \cdot \vec{1} = V \\ 
\end{split}
\end{equation}

To solve this problem in a differentiable way, an entropic regularization term $H(T)$ is introduced as defined by $H(T) = - \sum_{i=1}^{N} \sum_{j=1}^{N} T_{ij} \log T_{ij}$.
Then the original problem can be formulated as below:
\begin{equation}
\begin{split}
& L_{sml} = \mathop{min}\limits_{T} \langle C, T \rangle - \epsilon H(T) \\
& subject \ to \quad T\cdot \vec{1} = U, \quad T^\top \cdot \vec{1} = V \\
\end{split}
\end{equation}
where $\epsilon$ is the regularization coefficients which denote the smoothness of the transportation plan matrix $T$. In our model, $\epsilon$ is set to 0.0001 empirically.
The regularized form of the problem can be solved efficiently by Sinkhorn iteration \cite{cuturi2013sinkhorn} in a differentiable way.

Using SML for the regression of spherical distribution has two clear advantages. First, it makes the regression sensitive to the global geometry, thus effectively penalizing predicted activation that is far away from the ground truth distribution. Second, SML is smooth in training which enables stable optimization which is beneficial to the under-constraint problem in illumination prediction.

\begin{figure*}[t]
\centering
\includegraphics[width=1.0\linewidth]{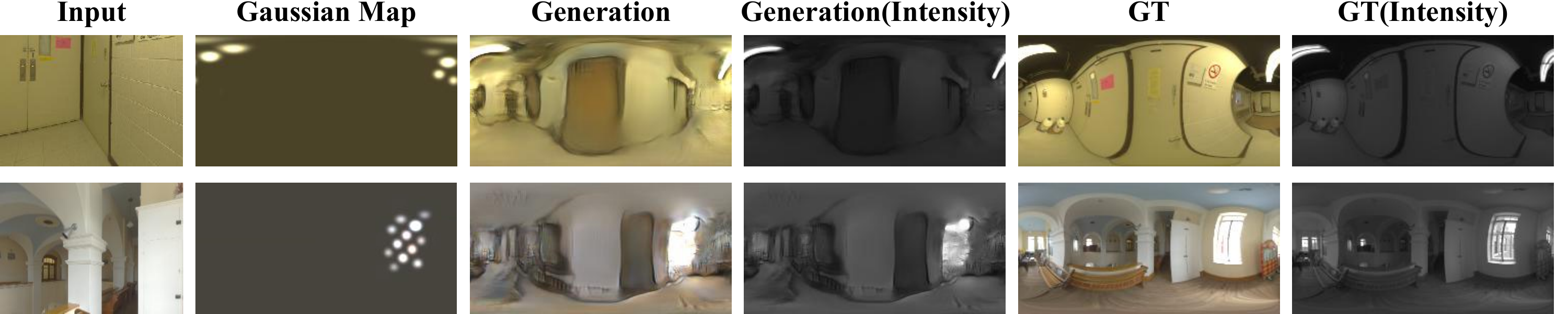}
\caption{Illustration of EMLight illumination estimation: For the input images in column 1, columns 2 shows the constructed Gaussian maps based on the regressed illumination parameters and columns 3 and 4 show the generated illumination map under the guidance of Gaussian map and the corresponding intensity map, respectively. Columns 5 and 6 show the ground truth of the illumination maps and the corresponding intensity maps, respectively.
}
\label{im_sample}
\end{figure*}

\begin{figure}[ht]
\centering
\includegraphics[width=1.0\linewidth]{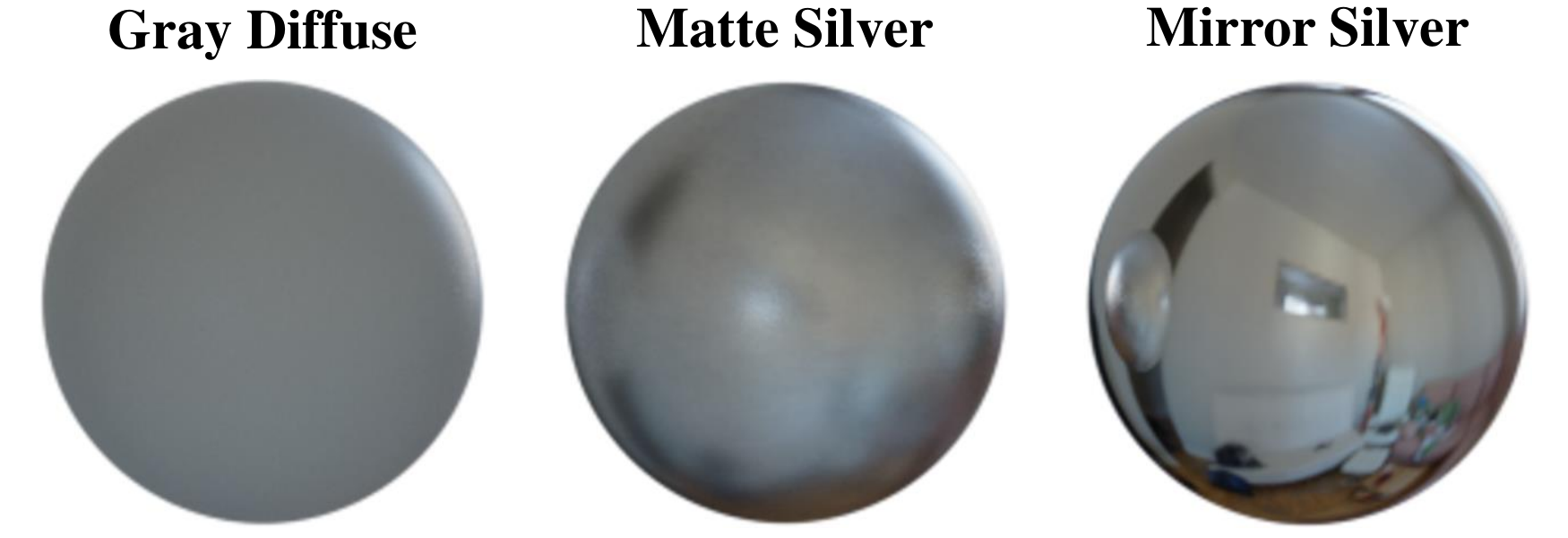}
\caption{
The scenes used in evaluations consist of three spheres with different materials including diffuse gray, matte silver and mirror silver.
}
\label{im_ball}
\end{figure}

\subsection{Neural Projector}

With the predicted light distribution, light intensity and ambient term, 
we propose a Neural Projector to formulate the synthesis of illumination map as a conditional image generation task with paired data as illustrated in Fig. \ref{im_stru2}.
To synthesize realistic frequency information in the illumination map, the neural projector is trained in an adversarial manner. 
The input to the neural projector includes the predicted illumination parameters and the input image.
Firstly, we map the parameters into a Gaussian map through spherical Gaussian function \cite{gardner2019deeppara} as shown below:
\begin{equation}
M = \sum_{i=1}^{N} v_i * exp \frac{d_{i}*u-1}{s} + A
\end{equation}
where $M$ is the gaussian map, $N$ is the number of anchor points, $v_{i}$ denotes the RGB value of a anchor points which is the product of light distribution on this anchor point and light intensity (namely $v_{i} = P_{i}*I$), $d_{i}$ is the direction of an anchor point (pre-defined by Vogel's method \cite{vogel1979}), $u$ is a unit vector giving a direction on the sphere, $s$ is the angular size (selecting 0.0025 empirically), $A$ is the ambient term.
Then the constructed Gaussian map will be treated as a guidance (or a condition) for the following generation.

The overall architecture of the generation part is similar to SPADE \cite{park2019spade} as shown in Fig. \ref{im_stru2}.
Instead of sampling a random vector, we encode the input image as a latent feature vector for the adversarial generation. 
The illumination map is panoramic image and pixels at different latitudes of a panorama are stretched in different scales. As a result, normal convolution suffers from distortions heavily at different latitudes especially around the polar regions of the panoramic image.
Previous work SphereNet \cite{spherenet} builds on regular convolutional filters, which naturally enables the transfer of CNNs between different image representations by adapting the sampling locations of the convolution kernels. 
We therefore adopt spherical convolution (SConv Block) for the generation of panoramic illumination map, effectively reversing distortions and wraps the filters around the sphere.
The Gaussian map is then fused with the feature of SConv Block in multiple scales through the spatially-adaptive normalization as described in \cite{park2019spade}.
The details of generation part is provided in the supplementary file.

The neural projector employs several losses to drive the generation of high-quality illumination maps. We denote the input Gaussian map as $x$, the ground-truth illumination map as $y$, and the generated illumination map as $x'$. To stabilize the training, we introduce a feature matching loss $\mathcal{L}_{feat}$ to match the intermediate features of discriminator between the generated illumination map and ground truth: 
\begin{equation}
    \mathcal{L}_{feat} = \sum_{l} \lambda_{l} ||D_{l}(x, x') - D_{l}(x, y) ||_{1}
\end{equation}
where $D_{l}$ represents the activation of layer $l$ in the discriminator and $\lambda_{l}$ denotes the balanced coefficients.
To obtain a similar illumination distribution instead of excessively emphasizing the absolute intensity,
a cosine similarity is computed between the generated illumination map and ground truth as below:
\begin{equation}
    \mathcal{L}_{cos} = (1 - Cos(x', y)) * \lambda_{cos}
\end{equation}
where $\lambda_{cos}$ is the weight of this term.
The discriminator adopts the same architecture with Patch-GAN  \cite{isola2017pixel2pixel}, thus obtaining the adversarial loss of discriminator $D$ and generator $G$ as denoted by $\mathcal{L}_{D}$ and $\mathcal{L}_{G}$.
Then the neural projector is optimized following the objective as below:
\begin{equation}
    \mathcal{L} = \mathop{min}\limits_{G} \mathop{max}\limits_{D} (\mathcal{L}_{feat} + \mathcal{L}_{cos} + \mathcal{L}_{G} + \mathcal{L}_{D})
\end{equation}
As the regression network and neural projector are all differentiable, the whole framework can be optimized end-to-end.

\renewcommand\arraystretch{1.25}
\begin{table*}[t]
\small 
\caption{
Comparison of EMLight with several state-of-the-art lighting estimation methods: The evaluation metrics include the widely used RMSE, si-RMSE, Angular Error and AMT. D, S, M denote a diffuse, a matte silver and a mirror material of the rendered objects, respectively.
}
\renewcommand\tabcolsep{3.5pt}
\centering 
\begin{tabular}{l|p{0.8cm}<{\centering} p{0.8cm}<{\centering} p{0.8cm}<{\centering} |
p{0.8cm}<{\centering} p{0.8cm}<{\centering} p{0.8cm}<{\centering} |
ccc|
p{0.75cm}<{\centering} p{0.75cm}<{\centering} p{0.75cm}<{\centering} |
p{0.75cm}<{\centering} p{0.75cm}<{\centering} p{0.75cm}<{\centering}
} \hline
\multirow{2}{*}{\textbf{Metrics}} & 
\multicolumn{3}{c|}{\textbf{\citet{gardner2017}}} & 
\multicolumn{3}{c|}{\textbf{\citet{gardner2019deeppara}}} & 
\multicolumn{3}{c|}{\textbf{\citet{li2019spherical}}} &
\multicolumn{3}{c|}{\textbf{\citet{garon2019fast}}} &
\multicolumn{3}{c}{\textbf{EMLight}} \\
\cline{2-16}
 & D & S & M & D & S & M& D & S & M & D & S & M & D & S & M \\\hline

\textbf{RMSE}    & 0.146 & 0.173 & 0.202   & 0.084 & 0.112 & 0.147   & 0.203 & 0.218 & 0.257   & 0.181 & 0.207 & 0.249    & \textbf{0.062} & \textbf{0.071} & \textbf{0.089}   \\

\textbf{si-RMSE} & 0.142 & 0.151 & 0.174   & 0.073 & 0.093 & 0.119   & 0.193 & 0.212 & 0.243   & 0.177 & 0.196 & 0.221    & \textbf{0.043} & \textbf{0.054} & \textbf{0.078}   \\

\textbf{Angular Error}  & 8.12$^{\circ}$ & 8.37$^{\circ}$ & 8.81$^{\circ}$   & 6.82$^{\circ}$ & 7.15$^{\circ}$ & 7.22$^{\circ}$   & 9.37$^{\circ}$ & 9.51$^{\circ}$ & 9.81$^{\circ}$  & 9.12$^{\circ}$ & 9.32$^{\circ}$ & 9.49$^{\circ}$   & \textbf{6.43$^{\circ}$} & \textbf{6.61$^{\circ}$} & \textbf{6.95$^{\circ}$}  \\

\textbf{AMT}     & 28.0\% & 23.0\% & 20.5\%   & 33.5\% & 28.0\% & 24.5\%   & 25.0\% & 21.5\% & 17.5\%   & 27.0\% & 22.5\% & 19.0\%    & \textbf{40.0\%} & \textbf{35.5\%} & \textbf{31.0\%}   \\
\hline

\end{tabular}
\label{tab_compare}
\end{table*}

\begin{figure*}[ht]
\centering
\includegraphics[width=1.0\linewidth]{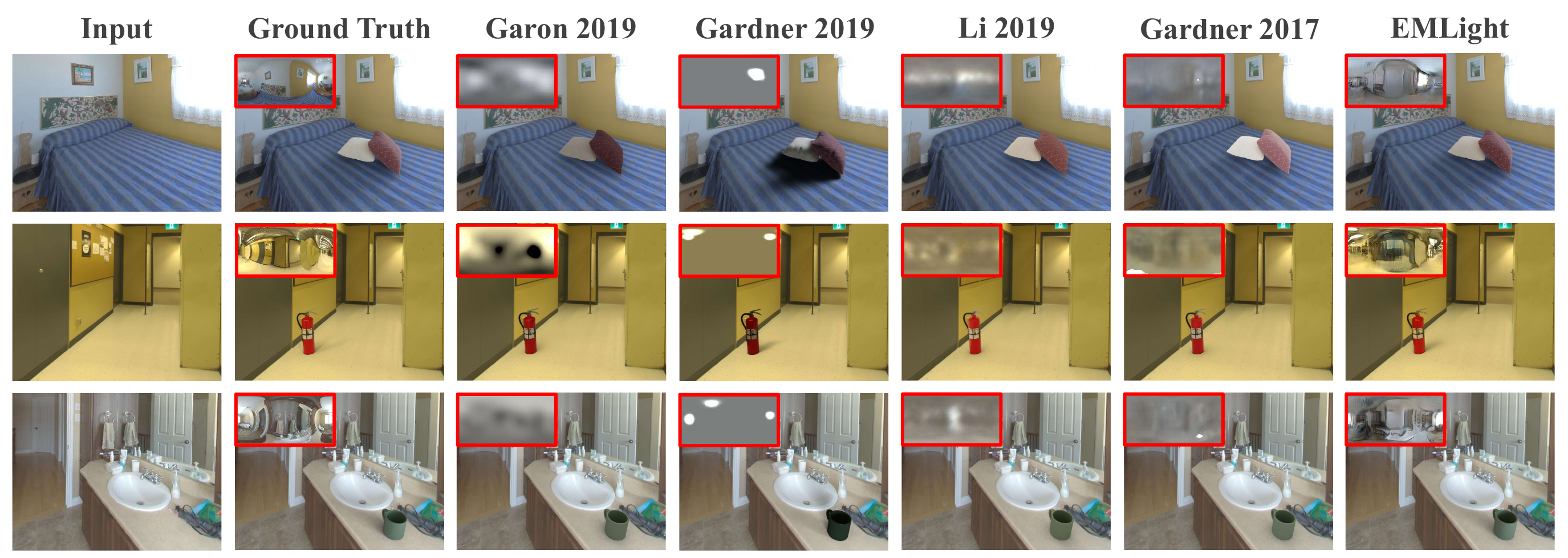}
\caption{
Visual comparison of EMLight with state-of-the-art lighting estimation methods: With the illumination maps predicted by different methods (at top-left corner of each rendered image), the rendered virtual objects demonstrate different light intensity, color, shadow and shade.
}
\label{im_render}
\end{figure*}

\section{Experiments}

\subsection{Dataset and Experimental Setting}
We evaluate EMLight with the Laval Indoor HDR Dataset \cite{gardner2017} that consists of 2,100 HDR panoramas taken in a variety of indoor environments. Similar to \citet{gardner2017}, we crop eight images with limited field of views from each panorama which produces 19,556 training pairs as used in our experiments. For each of the 19,556 images, the same image warping operation in \citet{gardner2017} is applied.
In the experiments, we randomly select 200 images as the testing set and the rest for training.

Consistent with \citet{gardner2019deeppara} and \citet{garon2019fast}, DenseNet121 is used as the backbone in regression network. 
Detailed network structure of neural projector and the training settings are provided in the supplementary file.

\subsection{Evaluation Method and Metric}

Similar to the evaluation setting in DeepLight \cite{legendre2019deeplight}, our scenes for evaluations include three spheres with different materials: gray diffuse, matte silver and mirror as illustrated in Fig. \ref{im_ball}. The performance is evaluated by comparing the scene images rendered (by Blender \cite{blender}) with predicted illumination maps and ground truth. 
The evaluation metrics include Root mean square error (RMSE) and scale-invariant RMSE (si-RMSE) that focus on the estimated light intensity and light directions (or shadings), respectively. Both metrics have been widely adopted in the evaluation of illumination prediction. In addition, we also adopt the per-pixel linear RGB angular error \cite{legendre2019deeplight} and Amazon Mechanical Turk (AMT) which performs crowdsourcing user study for subjective assessment of empirical realism of rendered images. In the experiments, each compared model predicts 200 illumination maps on the testing set for quantitative evaluation. For qualitative evaluation, we design 25 scenes for 3D insertion and render them with the predicted illumination maps.

\renewcommand\arraystretch{1.25}
\begin{table*}[t]
\small 
    \caption{Ablation study of the proposed EMLight: SG and SD denote the spherical Gaussian representation and our spherical distribution representation of the illumination map. 
    L2 and SML denote using the L2 and spherical mover's loss to regress the representation parameters.
    NP denotes the proposed neural projector.
}
\renewcommand\tabcolsep{5.75pt}
\centering 
\begin{tabular}{l||ccc||ccc||ccc||ccc} \hline
\multirow{2}{*}{\textbf{Models}} & 
\multicolumn{3}{c||}{\textbf{RMSE}} & 
\multicolumn{3}{c||}{\textbf{si-RMSE}} &
\multicolumn{3}{c||}{\textbf{Angular Error}} &
\multicolumn{3}{c}{\textbf{AMT}}
\\
\cline{2-13}
 & D & S & M & D & S & M& D & S & M &  D & S & M   \\\hline

\textbf{EMLight} (SG+L2) & 0.204 & 0.213 & 0.238   & 0.188 & 0.203 & 0.229     & 9.18$^{\circ}$  & 9.42$^{\circ}$  & 9.73$^{\circ}$   & 26.0\% & 22.5\% & 18.0\% \\

\textbf{EMLight} (SD+L2) & 0.133 & 0.161 & 0.178   & 0.117 & 0.132 & 0.161     & 7.60$^{\circ}$  & 7.88$^{\circ}$  & 8.12$^{\circ}$   & 30.5\% & 25.5\% & 22.0\%  \\

\textbf{EMLight} (SD+SML)     & 0.080 & 0.103 & 0.117   & 0.072 & 0.087 & 0.106      & 6.78$^{\circ}$  & 6.98$^{\circ}$  & 7.12$^{\circ}$   & 34.0\% & 31.5\% & 26.0\% \\

\hline
\textbf{EMLight} (SD+SML+NP)         
& \textbf{0.062} & \textbf{0.071} & \textbf{0.089}   
& \textbf{0.043} & \textbf{0.054} & \textbf{0.078}
& \textbf{6.43$^{\circ}$}  & \textbf{6.61$^{\circ}$} & \textbf{6.95$^{\circ}$}
& \textbf{40.0\%} & \textbf{35.5\%} & \textbf{31.0\%}    \\\hline
\end{tabular}
\label{tab_ablation}
\end{table*}


\renewcommand\arraystretch{1.3}
\begin{table}[t]
\caption{Ablation studies over anchor points, loss functions and convolution types: EMLight* denotes the standard EMLight with 128 anchor points, spherical mover's loss (SML), and spherical convolution. We create four EMLight variants by setting the number of anchor points to 64 and 196, replacing SML with cross-entropy loss, and replacing spherical convolution with normal convolution.
}
\renewcommand\tabcolsep{6.5pt}
\centering 
\begin{tabular}{l|p{1.5cm}<{\centering}|p{1.5cm}<{\centering}} \hline
\textbf{Models} & \textbf{RMSE} & \textbf{si-RMSE}
\\
\cline{2-3}
\hline\hline

\textbf{Anchor=64}       & 0.091  & 0.075    \\
\textbf{Anchor=196}      & 0.076  & \textbf{0.055}    \\

\hline

\textbf{Cross-Entropy Loss} & 0.102  & 0.082      \\

\hline

\textbf{Normal Convolution}  & 0.086 & 0.071 \\

\hline
\hline

\textbf{EMLight*}      & \textbf{0.074} & 0.058    \\\hline
\end{tabular}
\label{tab_ablation2}
\end{table}


\subsection{Quantitative Evaluation}
We compare EMLight with several state-of-the-art methods that estimate illumination maps directly \cite{gardner2017} or estimate representative illumination parameters \cite{garon2019fast,li2019spherical,gardner2019deeppara}. For each compared method, we render 200 images of the testing scene (three spheres with diffuse, matte silver, mirror silver materials) by using the illumination maps predicted from the testing set. Table \ref{tab_compare} shows experimental results, where \textit{D}, \textit{S} and \textit{M} denote diffuse, matte silver and mirror material of the objects to be rendered, respectively. AMT user study is conducted by showing two images rendered by the ground truth and each compared methods in Table \ref{tab_compare} to 20 users who will pick more realistic image. The score is the percentage of rendered images (200 images in total) that is deemed as more realistic than the ground-truth rendering.

We can observe that EMLight outperforms all compared methods under different evaluation metrics and materials consistently, largely attributed to the accurate generation of illumination under the guidance of the Gaussian map. \citet{gardner2017} predict illumination maps directly but the direct generation without any guidance tends to over-fit training data and lead to sub-optimal generalization due to the unconstrained nature of illumination estimation from a single image. \citet{gardner2019deeppara} regress spherical Gaussian parameters of light sources which tends to lose useful frequency information and generate inaccurate shading and shadow as measured by si-RMSE. \citet{li2019spherical} adopt spherical Gaussian functions to reconstruct the full illumination map in the spatial domain which often fails to recover high-frequency illumination. \citet{garon2019fast} recover lighting by regressing spherical harmonic coefficients while it struggles to regress accurate light directions and recover high-frequency information. Though a masked L2 loss is employed in \citet{garon2019fast} for preserving high-frequency information, it does not solve the problem essentially as illustrated in Fig. \ref{im_render}. As a comparison, EMLight estimates illumination parameters accurately by regressing light distribution under a spherical mover's loss. Under the guidance of estimated parameters, the neural projector generates accurate and high-fidelity illumination maps with realistic frequency information through adversarial training.

\subsection{Qualitative Evaluation}

We visualize our predicted Gaussian maps, generated illumination maps, and the corresponding intensity maps in Fig. \ref{im_sample}. As Fig. \ref{im_sample} shows, our regression network predicts light distribution accurately as shown in \textit{Gaussian Map}. The neural projector generates accurate and realistic HDR illumination maps as shown in \textit{Generation}.
To further verify the quality of generated HDR illumination, we visualize the intensity maps of the illumination maps.

We compare EMLight with four state-of-the-art light estimation methods qualitatively. Fig. \ref{im_render} shows rendered images with the predicted illumination maps (highlighted by red boxes). We can observe that EMLight predicts realistic illumination maps with plausible light sources and produces realistic rendering with clear and accurate shade and shadows.
As a comparison, direct generation \cite{gardner2017} struggles to identify the direction of light sources as there is no guidance for the generation.
Illumination maps by Gardner et al. \cite{gardner2019deeppara} are over-simplified with a limited number of light sources, and the simplification loses accurate frequency information which results in unrealistic shadow and shading in rendering. Garon et al. \cite{garon2019fast} and Li et al. \cite{li2019spherical} regress illumination parameters but are often constrained by the order of representative functions (spherical harmonic and spherical Gaussian). As a result, they predict illumination of low frequency and produce renderings with very weak shade and shadow in rendering as illustrated in Fig. \ref{im_render}.

\subsection{Ablation Study}

We develop several EMLight variants as listed in Table \ref{tab_ablation} to evaluate the effectiveness of proposed methods.
The variants include 1) the baseline \textit{EMLight (SG+L2)} that regresses spherical Gaussian representative parameters with L2 loss as in \cite{gardner2017}; 2) the \textit{EMLight (SD+L2)} that regresses the spherical distribution of illumination with L2 loss; 3) the (\textit{EMLight (SD+SML)}) that regresses the spherical distribution of illumination with SML; and 4) the standard model \textit{EMLight (SD+SML+NP)}. Similar to the setting in \textit{Quantitative Evaluation}, we apply all variant models to render 200 images of the testing scene. As Table \ref{tab_ablation} shows, (\textit{EMLight (SD+L2)}) outperforms \textit{EMLight (SG+L2)} clearly, demonstrating the superiority of the spherical distribution representation of illumination. \textit{EMLight(SD+SML)} also produces better estimation than \textit{EMLight(SD+L2)}, validating the effectiveness of SML.
\textit{EMLight (SD+SML+NP)} achieves the best estimation, demonstrating that the neural projector promotes the performance of illumination prediction significantly.

We also benchmark spherical mover's loss (SML) with the widely adopted Cross-Entropy Loss for distribution regression, compare the spherical convolution with normal convolution, and study how anchor points affect the light estimation as shown in Table \ref{tab_ablation2}. We followed the experimental setting as in Table \ref{tab_ablation} and measure the averaged RMSE and si-RMSE on three materials. As Table \ref{tab_ablation2} show, SML outperforms cross-entropy loss clearly as SML captures spatial information of spherical distributions effectively. In addition, spherical convolution performs better than normal convolution consistently in panoramic image generation. Further, the prediction performance drops slightly when 64 instead of 128 anchor points are used, and increasing anchor points to 196 doesn't improve the performance obviously. We conjecture that the larger number of parameters with 196 anchor points affects the regression accuracy negatively.

\section{Conclusions}
This paper presents EMLight, a lighting estimation framework that formulates the illumination prediction as a light distribution regression problem over a spherical surface. A spherical mover's loss is proposed to achieve the effective regression of spherical light distribution. To generate illumination maps with realistic frequency information, we introduce a novel neural projector with spherical convolution that generates panoramic illumination maps through adversarial training. Quantitative and qualitative experiments show that EMLight is capable of predicting illumination accurately from a single indoor image. We will continue to investigate illumination estimation from the perspective of spherical distributions in our future works.


\bibliography{cite}
\end{document}